\pdfoutput=1

\documentclass[11pt]{article}

\usepackage[final]{coling}

\usepackage{times}
\usepackage{latexsym}
\usepackage{amssymb}
\usepackage{multirow}
\usepackage{placeins}
\usepackage{longtable}

\usepackage{xurl} 

\usepackage[T1]{fontenc}

\usepackage[utf8]{inputenc}

\usepackage{microtype}

\usepackage{inconsolata}

\usepackage{graphicx}

\usepackage{pifont}%
\newcommand{\xmark}{\ding{55}}%

\title{Recent Advancements and Challenges of Turkic Central Asian Language Processing}

\author{Yana Veitsman, Mareike Hartmann \\
  Department of Language Science and Technology \\ 
  Saarland University, Germany \\
  \texttt{\{yanav, mareikeh\}@lst.uni-saarland.de} \\}

\begin{document}
\maketitle
\begin{abstract}

Research in NLP for Central Asian Turkic languages - Kazakh, Uzbek, Kyrgyz, and Turkmen - faces typical low-resource language challenges like data scarcity, limited linguistic resources and technology development. However, recent advancements have included the collection of language-specific datasets and the development of models for downstream tasks. Thus, this paper aims to summarize recent progress and identify future research directions. It provides a high-level overview of each language's linguistic features, the current technology landscape, the application of transfer learning from higher-resource languages, and the availability of labeled and unlabeled data. By outlining the current state, we hope to inspire and facilitate future research.
\end{abstract}

\section{Introduction}
\begin{figure}[ht!]
    \centering
    \includegraphics[width=1\linewidth]{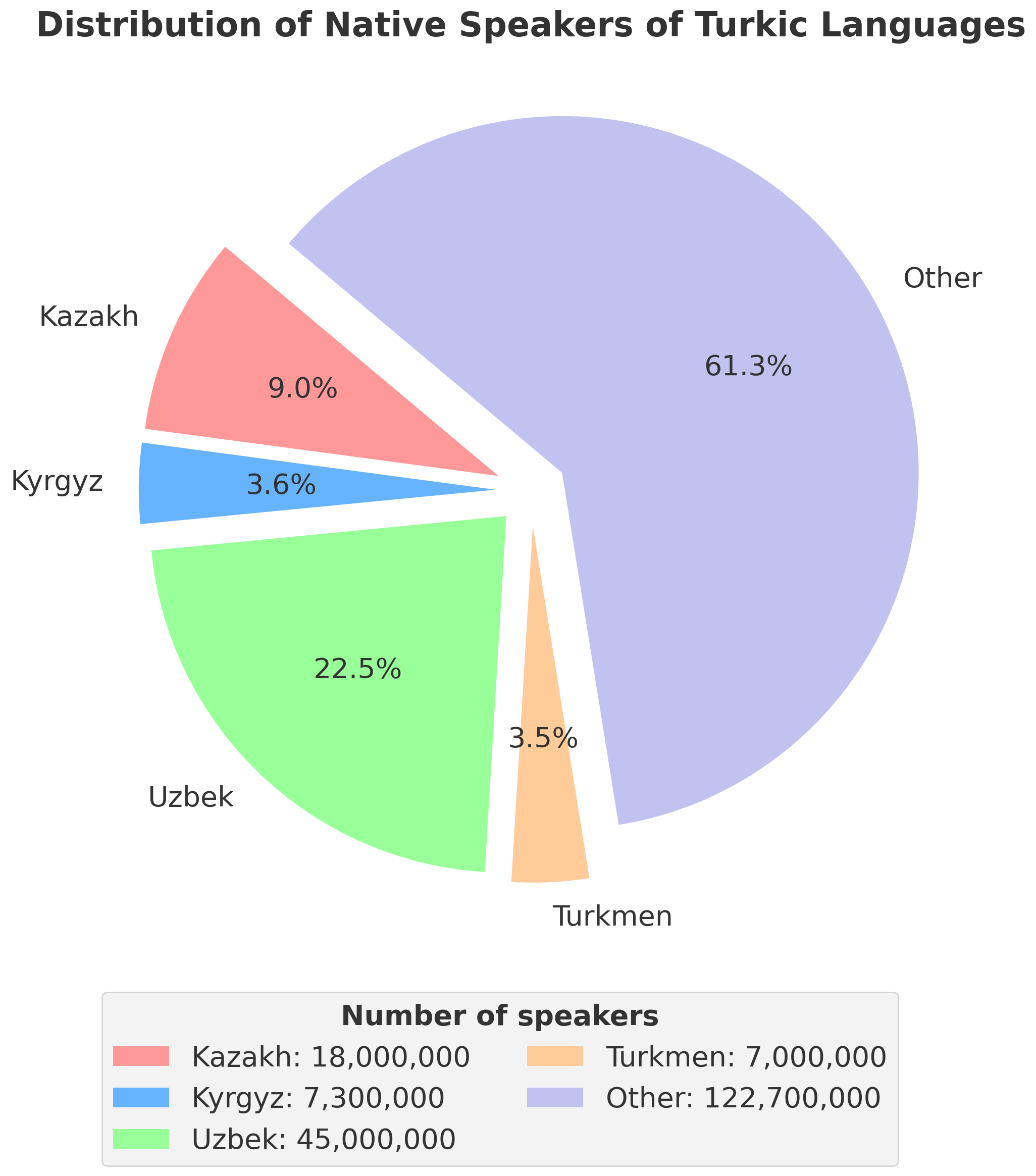}
    \caption{Distributions of numbers of Kazakh, Uzbek, Kyrgyz, and Turkmen native speakers among all Turkic language speakers. Numbers in the legend are approximates. Source: \url{https://en.wikipedia.org/w/index.php?title=Languages_of_Asia&oldid=1230214231}}
    \label{fig:speakers-distr}
\end{figure}

Turkic languages are spoken by approximately 200 million people worldwide, with a significant concentration in Central Asia (see detailed breakdown of the number of speakers in Figure \ref{fig:speakers-distr}). While Turkish is the most resourceful language in the family, this paper focuses on less-resourced Turkic languages that are geographically, historically, and linguistically closer to one another in Central Asia. These languages represent an important subset of the Turkic family, spoken by approximately 80 million people in the region. \\
Like any other low-resource language speakers, speakers of Central Asian languages would benefit from having reliable language technology, from simple spell checkers to virtual assistants. Such tools would uphold newly adopted language policies and cement the role of local languages in the region. Developing these resources, however, requires the existence of open-source datasets and up-to-date language models. To address these resource limitations, researchers are exploring methods like transfer learning and data augmentation, though both have limitations in task applicability and effectiveness \cite{chen2021empiricalsurveydataaugmentation, Raffel2019ExploringTL}. \\
This paper aims to provide an overview of existing resources and suggest directions for future research to support both those utilizing current resources and those developing new ones (for the search strategy details see Appendix \ref{appendix:search}). We also seek to highlight current resource needs, addressing which of those could be particularly crucial in advancing the Turkic Central Asian languages toward a higher-resource status.

\section{Related Work}
Recently, substantial efforts have been made to consolidate domain knowledge for Turkic languages via linguistic analysis tools \cite{akin2007zemberek, 10017049}, and NLP technology assessments \cite{mirzakhalov-etal-2021-large, maxutov-etal-2024-llms}. However, there is still a lack of comprehensive research summarizing the available data and language processing tools, especially for Central Asian Turkic languages. While state-of-the-art advancements in speech recognition and machine translation exist for some languages \cite{end2endASR,yeshpanov2024kazparc}, no cross-linguistic comparisons have been conducted. A detailed survey could provide a valuable foundation for comparison and help define future research directions.

\begin{table*}[t]
\centering
    \begin{tabular}{l  c  c  c  c  c}
    \hline
         \textbf{Feature} & \textbf{Turkish} & \textbf{Kazakh} & \textbf{Kyrgyz} & \textbf{Uzbek} & \textbf{Turkmen}\\
         \hline
         Number of vowels & 8 & 12 & 8 & 6 & 9\\
         Number of plural suffixes & 2 & 12 & 12 & 4 & 4 \\
         Number of pronouns & 6 & 8 & 8 & 8 & 6\\
         Number of noun cases & 6 & 7 & 6 & 6 & 6\\
         Number of personal verb suffixes & 5 & 9 & 6 & 9 & 5\\
         Word order & SOV & SOV & SOV & SOV & SOV\\
         \hline
    \end{tabular}
    \caption{High-level overview of the Turkic Central Asian languages' differences. Sources: \url{https://ecosystem.education/doc/Turkic\%20Diller-SS.pdf}, \url{https://www.britannica.com/topic/Turkic-languages}}
    \label{tab:ling_features_per_lang}
\end{table*}

\section{Difficulties in Processing Turkic Languages}
\subsection{Overview}
Typology has a potential to improve language processing and transfer learning \cite{ponti-etal-2019-modeling}, and learning from similar languages is overall beneficial for the latter \cite{zoph-etal-2016-transfer}. Therefore, providing an overview of the linguistic features of Turkic languages may help identify potential similarities and challenges in their processing.  While in depth linguistic comparison of each language with their richer counterpart, Turkish \cite{_ltekin_2022}, lies beyond the scope of the paper, a basic summary is provided in \autoref{tab:ling_features_per_lang}. \\
Turkic languages are primarily described as morphologically rich. Over the past years, several studies revolving around the specifics of processing their complex morphological structures and morpho-syntactic features have been conducted \cite{Ciddi_2013}. These studies showed that the highly agglutinative nature of Turkic languages is specifically problematic in machine translation \cite{Alkim2019MachineTI, mirzakhalov-etal-2021-large} and named entity recognition \cite{Kk2017NamedER} tasks. Additionally, it is not exactly clear how these peculiarities will be reflected in the transfer learning applications.

\subsection{Similarities and Differences}
Many grammatical features that impact natural language processing - such as word order and verb tense systems - are common across all the languages considered. Word order similarity is particularly crucial because of its impact on multilingual learning \cite{dufter-schutze-2020-identifying}. On the other hand, shared verb tense systems can facilitate cross-lingual data enrichment \cite{asgari-schutze-2017-past}, important for transfer learning. Nevertheless, minor differences among the languages do exist, including a heavier reliance on vowel harmony in Kazakh, Uzbek, Kyrgyz, and Turkmen. This results in additional noun cases and plural suffixes, with the exact form depending on the vowel of the preceding syllable. \\ 
Analyzing the linguistic differences in \autoref{tab:ling_features_per_lang}, such as variations in the number of pronouns or vowel sounds, one can infer that Kazakh and Kyrgyz are typologically closer to each other than Kazakh is to Uzbek, or Uzbek is to Kyrgyz; similarly, Turkmen is grammatically closest to Turkish. This suggests not only potential differences in transfer learning efficacy from Turkish, but also the possibility of successful transfer learning within the Central Asian group itself - likely making transfer between certain pairs, such as Kazakh and Kyrgyz, more effective than between others. \\
Another significant difference among these languages is the script used and its inconsistent application. For example, while Uzbek is written in Latin script, Kazakh, despite efforts by the Kazakhstani government to adopt Latin script, is still primarily written in Cyrillic. Such script inconsistencies could limit the effectiveness of transfer learning \cite{gheini2019universalparentmodellowresource, amrhein-sennrich-2020-romanization}, suggesting that additional steps in data cleaning or preprocessing may be necessary to optimize model performance.\\
Overall, these observations offer a promising opportunity to leverage linguistic similarities for transfer learning both from Turkish and Turkic Central Asian languages themselves.

\section{Datasets Availability}
Open-source language resources enable researchers to scale and reuse data, reducing overall time spent on corpora collection and evaluation. However, data access varies from low to almost non-existent for the languages considered in the paper. A full summary of all the reviewed datasets and their primary features can be found in the Appendix~\ref{appendix:datasets}.

\subsection{Sources of Data and Stakeholders}
While developing a comprehensive dataset classification system would lie beyond the scope of the current paper, it is important to identify potential data collection sources and main stakeholders of the process. While a significant effort has been made by the researchers within the Central Asian countries themselves (e.g., most notable research on the Kazakh language is contributed by the Institute of Smart and Intelligent Systems (ISSAI) of Nazarbayev University), outside effort is driven primarily by the studies of NLP applications for Chinese minority languages \cite{app13010326} or Turkic languages in general \cite{mirzakhalov-etal-2021-large}. Additionally, while the researchers focus on collection of handcrafted high-quality datasets, other, potentially lower-quality data, is available via multilingual datasets crawled from the web, such as Common Crawl (CC)\footnote{\url{https://www.commoncrawl.org}} or OSCAR \cite{ortiz-suarez-etal-2020-monolingual}. One more potential data source is the machine-translated content from high-resource languages; however, assessing the exact quantity of such data poses to be challenging, so the survey does not focus on it. Thus, there is a greater variety of data sources and stakeholders than might initially appear. 

\subsection{Kazakh Language Datasets}
\textbf{Unannotated Datasets and Corpora Collections.} One of the largest available dataset of unannoted text articles in Kazakh is the Kazakh Language Corpus \cite{makhambetov-etal-2013-assembling}, containing a broad range of written text on a variety of topics. Additionally, annotated sub-corpora with linguistics and other language features are available within the same dataset. Another large collection of texts is the Kazakh National Language Corpus\footnote{\url{https://qazcorpora.kz}} with over 22 thousand documents available.\\
\textbf{Linguistic Features Datasets.} Among the four surveyed languages, Kazakh has the most diverse and extensive datasets available. Besides collections of morphologically and syntactically-annotated data like Almaty Corpus of Kazakh Language\footnote{\url{http://web-corpora.net/KazakhCorpus/search/?interface_language=en}} and UD treebank \cite{tyers_tl2015, makazhan_tl2015}, there exist also more task-specific text corpora. \\
\textbf{Task-specific Datasets.} For example, the KazNERD corpus, spanning over 100,000 sentences with 25 entity classes for the task of Named Entity Recognition \cite{yeshpanov-etal-2022-kaznerd}, the KazQAD corpus for open-domain question answering with over 6000 questions \cite{yeshpanov2024kazqad}, the KazParC parallel corpus of Kazakh, English, Russian, and Turkish for machine translation \cite{yeshpanov2024kazparc}, and the KazSAnDRA dataset, containing over 180 thousand reviews to be used for the task of sentiment analysis \citep{yeshpanov2024kazsandra}, are all publicly available. \\
\textbf{Multimodal Datasets.} There are also numerous collections of multimodal data for the language. Most notable and largest language dataset up until now is the Kazakh Speech Corpus 2 (KSC2), collected by \citet{Mussakhojayeva2022KSC2AI}, which contains over 1,200 hours of transcribed audios, and which also integrates research work on previously collected datasets, such as Kazakh Speech Corpus (KSC), compiled by \citet{khassanov2021crowdsourced}. Another prominent resource is the KazakhTTS2 dataset \cite{mussakhojayeva-etal-2022-kazakhtts2}, which extended its previous version, KazakhTTS \cite{Mussakhojayeva2021KazakhTTSAO}, to more than 270 hours of recorded audio. Drawing from this work on speech, a speech commands dataset was also recently collected \cite{speech-commands-kaz}. Additionally, more multimodal data is available in Kazakh for even more peculiar applications; for example, \citet{Mukushev2022TowardsLV} have gathered over 40,000 video samples of 50 signers of Kazakh-Russian Sign Language, which was also an expansion of a previous project by the same authors in 2020. Further work on the topic includes recent development of the KazEmoTTS dataset \cite{abilbekov2024kazemotts} that collects more than 54 thousands audio-text pairs in over 5 emotional states. Several datasets were also collected for the purpose of hand-written text recognition, some of them combining Kazakh and Russian languages \cite{Abdallah_2020, hkr}, and some containing text inclusively in Kazakh, for example, Kazakh Offline Handwritten Text Dataset \cite{Toiganbayeva_2022}, with over 3000 exam papers available. \\
This variety of multimodal data can potentially cover a wide range of the data needs, enabling efficient data reuse and eliminating the necessity to use precious resources for additional data collection. For example, the audios from the speech corpus can be transcribed and used as text for the tasks of text classification, text generation, information retrieval, and others. Thus, these datasets have a great potential of making a strong contribution in development of other NLP tasks.

\subsection{Uzbek Language Datasets}
Uzbek ranks second in terms of data availability among Central Asian languages; however, data that exclusively covers this language remains relatively scarce comparing even to the Kazakh language alone. \\
\textbf{Linguistic Features Datasets.} A distinguishing feature of most Uzbek datasets is their focus on purely linguistic tasks; for example, several datasets, such as UzWordnet \cite{agostini-etal-2021-uzwordnet} and SimRelUz \cite{salaev2022simreluz} capture prominent semantic features of the language; others, such as a dataset collected by \citet{Sharipov2023UzbekTaggerTR}, that includes a variety of POS tags and syntactic features of this Central Asian language.\\
\textbf{Task-specific Datasets.} There is substantial data for certain tasks, such as sentiment analysis. For example, there exist a sentiment analysis dataset collected by \citet{Kuriyozov2019BuildingAN} and one based on restaurant reviews by \citet{matlatipov2022uzbek}. There also exists data for text classification, mainly scraped from the news sources in Uzbek, one collected by \citet{Rabbimov2020MultiClassTC} and another one by \citet{kuriyozov2023text}.\\
\textbf{Other Datasets.} Some additional datasets, including multimodal and/or general unannotated data also exist. For example, one of the multimodal datasets is the open-source Uzbek Speech Corpus \cite{musaev2021usc}. On the other hand, there is the Uzbek corpus\footnote{\url{https://uzbekcorpus.uz/}}, which includes a large collection of educational, scientific, official, and artistic texts together with a morphological database and dictionaries, and an Uzbek Community corpus\footnote{\url{https://corpora.uni-leipzig.de/en?corpusId=uzb_community_2017}}, collected by Leipzig University and containing over 660 thousand sentences of community data only. 

\subsection{Kyrgyz Language Datasets}
\textbf{Unannotated Datasets and Corpora Collections.} One of the largest open-source datasets for Kyrgyz is the Manas-UdS corpus\footnote{\url{https://fedora.clarin-d.uni-saarland.de/kyrgyz/index.html}} of over 84 literary texts in 5 genres, marked with lemmas and parts of speech \cite{kasieva-kyrgyz}. For most other datasets, linking them directly to publications is not possible; however, some of them are available on Github. One of those is the KyrgyzNews dataset\footnote{\url{https://github.com/Akyl-AI/Kyrgyz_News_Corpus}} with over 250 thousand scraped news. Leipzig University has also compiled a corpus of over 3 million sentences from publicly available web sources.\footnote{\url{https://corpora.uni-leipzig.de/de?corpusId=kir_news_2020}}\\
\textbf{Task-specific datasets.} The few available task-specific datasets also can only be found on Github. One such example, for the task of NER, is the NER dataset\footnote{\url{https://github.com/Akyl-AI/KyrgyzNER}} that is currently under development by the researchers from Kyrgyz State Technical University (KSTU). Another example is the hand-written letters dataset\footnote{\url{https://github.com/Akyl-AI/kyrgyz_MNIST}} (Kyrgyz MNIST equivalent) also available on Github.

\subsection{Turkmen Language Datasets}
The data landscape of Turkmen language is even more scarce. Besides the Leipzig University’s corpora\footnote{\url{https://corpora.wortschatz-leipzig.de/en?corpusId=tuk-tm_web_2019}} of over 270 thousand sentences of web-scraped data, other resources are practically non-existent. With only a few dictionaries and poetry and literature collections\footnote{\url{https://github.com/tmLang-NLP/datasets}} collected by enthusiasts and available in an open-source fashion on Github, Turkmen language falls largely behind every other Turkic language of Central Asia in terms of data availability.

\subsection{Web-Scraped Datasets}
All the above-mentioned languages are also represented in multilingual datasets predominantly scraped from the web. In particular, all the languages are available in CC100 \cite{wenzek-etal-2020-ccnet}, WikiAnn \cite{pan-etal-2017-cross}, and OSCAR \cite{ortiz-suarez-etal-2020-monolingual} datasets. Most of other web-scraped datasets present online are either subsets of the bigger multilingual datasets (like CC100 or OSCAR), or are scrapes from newspapers and websites available on the Internet, with many of them not being open-source (for example, the kkWaC dataset\footnote{\url{https://www.sketchengine.eu/kkwac-kazakh-corpus/}} with over 139 million Kazakh words). However, as noted by \citet{Kreutzer_2022}, such datasets should be used with caution, as the quality of low-resource language data in them may be significantly lower. With only a few hundred labeled instances, even a 5-10 percent rate of mislabeled or grammatically incorrect sentences can substantially impact model performance in these languages.

\subsection{Multilingual Datasets}
Another important source of data in Central Asian languages is handpicked multilingual datasets for Turkic languages. For example, a large corpus was collected by \citet{Baisa2012LargeCF} for training morphological analyzers and disambiguators. Another example is the xSID (Cross-lingual Slot and Intent Detection) dataset by \citet{van-der-goot-etal-2020-cross}, which includes Turkic languages among other language families. More task-specific multilingual datasets featuring Central Asian languages include the Common Voice dataset \citep{ardila-etal-2020-common}, the Belebele dataset for machine reading comprehension containing Kazakh, Uzbek, and Kyrgyz \citep{bandarkar2023belebele}, the MuMiN dataset (Multimodal Fact-Checked Misinformation Dataset) based on the scraped tweets featuring Kazakh language \citep{nielsen2022mumin}, and the AM2iCo dataset for the evaluation of word meaning in context \citep{liu-etal-2021-am2ico}. Thus, Kazakh, Uzbek, and Kyrgyz make a prominent appearance in both the web domain and Turkic languages related research.

\subsection{Parallel Corpora}
Additionally, we would like to comment on the availability of parallel corpora for the languages studied. While OPUS provides a comprehensive overview of parallel data available \cite{tiedemann-thottingal-2020-opus} as well as several models \cite{tiedemann2023democratizing}, only a few less resourceful parallel pairs datasets exist, including the already mentioned Kazakh-Russian sign language corpora, Uzbek-Kazakh \citep{ALLABERDIEV2024110194} and Kazakh-Russian \citep{c3043905871741b6a9280441a90eb75d} parallel corpora for MT and ASR.

\subsection{Classifying Languages by Data Availability}
The resources available for each language are far from abundant. If one were to categorize them according to the classification suggested by \citet{joshi-etal-2020-state}, Kazakh would most likely fit into “The Rising Star” category, with its substantial presence in the web and good variety of multimodal datasets. However, the research on this language is pushed back by a lack of labeled data for downstream tasks. Uzbek, given its recent developments and greater variability in terms of the linguistic resources available, would be categorized as “The Hopeful,” given that in the next years the efforts for collecting the datasets will not fade. Kyrgyz and Turkmen, unfortunately, not being sufficiently backed up by streamlined research efforts, would be classified as “The Scraping-Bys,” with the future of their data collection processes yet unclear.\\

\section{Reasons of Data Scarcity}
Data scarcity in Central Asian languages stems from the widespread use of Russian, limited internet access, and a lack of AI-focused educational and technological initiatives.

\subsection{Russian Language in Central Asia}
During the Soviet era, Russian was the dominant language across all republics. National government since then have promoted national languages, but Russian remains influential in science, education, and politics \cite{Fierman2012Russian}. Russian media and online resources are widely accessible in Central Asia, facilitating intercultural communication but limiting the development of NLP for local languages. Heavy reliance on Russian sources for web-scraped data and media results in limited Kazakh, Uzbek, Kyrgyz, and Turkmen content online, restricting data diversity for these languages.

\subsection{Internet Access}
Limited Internet access is the second reason for data scarcity in the region. As stated before, while a substantial amount of data has been gathered by NLP researchers from the Internet based sources, only 38 and 21 percent of users in Kyrgyzstan and Turkmenistan respectively have access to the global net.\footnote{\url{https://blogs.worldbank.org/en/europeandcentralasia/how-central-asia-can-ensure-it-doesnt-miss-out-digital-future}} While the situation is somewhat better in Uzbekistan (with 55 percent of population being able to access the medium) and significantly better in Kazakhstan (around 79 percent of population), limited connectivity hinders users’ abilities to contribute to open-source encyclopedias, blogs, news sites, and more. Additionally, the lack of digitalized resources, such as electronic books, journals, audio transcripts, and video recordings, restricts information sharing within the region.

\subsection{Lack of Initiatives}
Being a demanding and resource-greedy field, natural language processing requires substantial and long-term financial investments. However, only a few exclusively AI-dedicated initiatives have been launched by the governments: for example, ISSAI (Institute of Smart Systems and Artificial Intelligence) in Kazakhstan or High Technology Park in Kyrgyzstan. However, these institutions are not solely dedicated to NLP research, and cover a wide range of topics in the AI domain, including robotics, IoT, computer vision, and others. Consequently, without a tailored initiative, it is difficult for the researchers to specialize in NLP-related research only.

\section{Application of Transfer Learning}
The situation when one of the languages in a language family is more resourceful than others is not unique to Turkic languages. This opens the door for potential transfer learning from Turkish or one of the Central Asian languages to the other languages within the same family. Usually, this is more attainable than collecting large datasets from scratch. \\
Transfer learning has been substantially studied in the domain of machine translation, and the choice of parent language has been highlighted as an important criteria for the technique application \cite{zoph-etal-2016-transfer}. Combining transliteration and byte-pair encoding, \citet{nguyen2017transfer} proved that this transfer learning approach might be suitable beyond high-resource to low-resource pairs, extending to pairs within low-resource only, especially the ones belonging to the agglutinative languages. The greater availability of NLP tools for Kazakh made it possible to assess transfer learning potential for some Turkic laguages lying beyond the scope of this paper, namely, Tatar \cite{tatar-tf}. Other studies on transfer learning from Kazakh have been conducted on the task of ASR \cite{Orel2023SpeechRF}. Some of them also aimed at using Russian as a source language, but the efforts proved to be less successful \cite{n2020development}.

\section{Data Augmentation, Transliteration, and R-Drop Regularization}
Apart from transfer learning, data augmentation techniques, R-Drop regularization, and transliterations have been substantially researched for enhancing model performance in some Turkic languages. \\
A study on the topic of sentence augmentation using large language models for Kazakh \cite{Bimagambetova2023EvaluatingLL} demonstrated that data augmentation works well for already resourceful languages and does so less successfully in the low-resource domain, which might seem somewhat obvious. However, practical applications using other data augmentation techniques, including phrase replacement, proved to significantly improve the BLEU score of certain language pairs, for example, Kazakh-Chinese \cite{Wu2023KazakhChineseNM}. Data augmentation with R-Drop regularization has also proved useful for the same Kazakh-Chinese translation task in other studies \cite{app131910589}.\\
Other techniques, such as dropout and transliteration, are often applied alongside transfer learning and data augmentation. Furthermore, multilingual models may outperform those using only transfer learning on certain tasks \cite{Nugumanova2022SentimentAO}. Altogether, these various techniques allow researchers to experiment with Central Asian language processing without requiring extensive data collection.

\begin{table*}[t]
\centering
    \begin{tabular}{l  c  c  c  c}
    \hline
         \textbf{Task} & \textbf{Kazakh} & \textbf{Uzbek} & \textbf{Kyrgyz} & \textbf{Turkmen}\\
         \hline
         Automatic Speech Recognition& \checkmark & \checkmark & \checkmark & \xmark\\
         Machine Translation & \checkmark & \checkmark & \checkmark & \checkmark \\
         Named-Entity Recognition & \checkmark & \checkmark & \checkmark & \xmark\\
         Text Generation & \checkmark & \xmark & \xmark & \xmark\\
         Sentiment Analysis & \checkmark & \checkmark & \xmark & \xmark\\
         Text Classification & \checkmark & \checkmark & \checkmark & \xmark\\
         POS Tagging & \checkmark & \checkmark & \checkmark & \xmark\\
         Text Summarization & \checkmark & \checkmark & \xmark & \xmark\\
         Question Answering & \checkmark & \xmark & \xmark & \xmark\\
         \hline
    \end{tabular}
    \caption{Existence of downstream task research per language. Existence is defined as at least one published research paper and/or dataset for the specific language and/or the specific language in conjunction with other closely related languages.}
    \label{tab:tech_per_lang}
\end{table*}

\section{Current State of Technologies}
\subsection{Kazakh Language Technologies}
In terms of available technology, Kazakh ranks first, just as it does in data availability. \\
\textbf{Linguistic Analysis and Rule-Based Systems.} A variety of tools have been developed for linguistic analysis and normalization of texts in Kazakh \cite{kaznlp}. Early on, rule-based translation systems and morphological analyzers have also been developed \cite{forcada-tyers-2016-apertium}. This research paved the way for the first advances in the most prominent areas of NLP, like machine translation and ASR, that are well represented in Kazakh language. \\
\textbf{Machine Translation.} Almost all the datasets mentioned in the previous sections have been released together with the evaluation benchmarks for certain tasks on either already pre-trained models like mBERT \cite{yeshpanov-etal-2022-kaznerd} or together with completely new systems like Tilmash, which enables a two-way translation between for 4 languages, including Kazakh and Turkish \cite{yeshpanov2024kazparc}. The latter has proved to be comparable or even better at translating language pairs involving Kazakh than translation technologies developed by Google and Yandex, which dominate the scene of machine translation in the region. \\
\textbf{ASR.} Another important advancement in processing of Kazakh language lies in the sphere of automatic speech recognition. For that purpose, researchers have been actively leveraging the KazakhTTS and KazakhTTS2 datasets as well as the recently available KazEmoTTS dataset. For example, a Turkic ASR system that employs the KazakhTTS, USC, and Common Voice data and covers, among other Turkic languages, Kazakh, Uzbek, and Kyrgyz, has been developed by the authors of the KazakhTTS dataset \cite{mussakhojayeva-etal-2022-kazakhtts2}. The most recent research on exclusively Kazakh language managed to bring down the word error rate to just 7.2 percent \cite{Bekarystankyzy2023TRANSFERLF} as well as to summarize difficulties and explore deep learning techniques on end-to-end speech recognition of agglutinative languages \cite{end2endASR}. Additionally, research on speech commands recognition has been conducted \cite{speech-commands-kaz}. \\
\textbf{Fine-tuning and Assessing Existing Models.} Regarding the usage and fine-tuning of already existing models, several tasks have been evaluated for Kazakh. For example, on the KazNERD dataset \cite{yeshpanov-etal-2022-kaznerd}, the fine-tuned XLM-RoBERTa demonstrated a micro average precision score of an impressive 97.09 percent, and on the task of sentiment analysis the same model gained an F1 score of 0.87. Additionally, \citet{maxutov-etal-2024-llms} assess the capabilities of 7 LLMs, including GPT-4 and Llama-2, on a variety of tasks, from classification to question answering, concluding that the performance of the models, just as expected. is lower on Kazakh language tasks in comparison to that on English language ones.\\
\textbf{Understudied Areas.} However, despite the above-mentioned efforts, certain areas of Kazakh NLP remain significantly understudied. One such example is the particularly dynamic field of text generation. Only a few experiments have been conducted regarding the usage of large language models for the benefit of Kazakh language \cite{Tolegen2023GenerativePT, maxutov-etal-2024-llms}.

\subsection{Uzbek Language Technologies}
Similar to data availability, Uzbek ranks second in terms of available technology. \\
\textbf{ASR.} While Uzbek does not enjoy the same variety of machine translation tools as Kazakh, presence of automatic speech recognition technology is somewhat comparable with the most recent work contributed by \citet{musaev2021usc}, with their model performing at the level of 14.3\% word error rate. \\
\textbf{Fine-Tuning Existing Models.} Uzbek also enjoys a better variety of pre-trained and fine-tuned models, including UzBERT \cite{mansurov2021uzbert}, capable of outperforming mBERT by leveraging at least 11 times more language specific data; UzRoberta, pre-trained on roughly 2 million news articles \cite{10.1063/5.0199871}; BERTBek, a model improving on UzBERT by training on Latin script \cite{kuriyozov-etal-2024-bertbek}, and a compact and fine-tuned variation of a mT5 model\footnote{\url{https://ijdt.uz/index.php/ijdt/article/view/104}}. In general, with Uzbek still catching up on data availability, there is potential for the language to soon enjoy a better variety of NLP technologies. 

\subsection{Kyrgyz and Turkmen Languages Technology}
Unfortunately, the situation for Kyrgyz and Turkmen looks less promising. With little to no work in terms of machine translation or automatic speech recognition, Kyrgyz enjoys little variety of technology for text classification \cite{Alekseev2023Benchmarking} and NER fine-tuned on WikiAnn data\footnote{\url{https://huggingface.co/murat/kyrgyz_language_NER}}. Turkmen offers even less language-specific technology, since it is has been mainly studied within the scope of comparative studies of Turkic languages. For example, some work on machine translation evaluation was done in the effort to build the infrastructure for Turkic languages, but nowadays this approach seems outdated \citep{Alkim2019MachineTI}. Overall, neither fine-tuning or pre-training of already available models and architectures have been researched for the majority of basic NLP tasks both in Kyrgyz and Turkmen.

\section{Future Work Areas}
\textbf{Kazakh Language.} Due to its variety of multimodal data, Kazakh language can potentially expand on the existing work and move onto fine-tuning and developing more advanced models for other downstream tasks, e.g. text generation or question answering. Additionally, transfer learning from Kazakh should be further explored as a potential workaround for the problems of its less resourceful counterparts, with a bigger focus on linguistically closer languages of the Turkic family, like Kyrgyz.\\
\textbf{Uzbek Language.} In contrast, Uzbek language clearly requires more data to be collected for the creation of the systems like the ones built on top of datasets like KazakhTTS. With already developed pre-trained models like UzBERT, there is a need for research into their further application and comparison with other models available. Generally, leveraging the rich Uzbek linguistic features datasets in combination with further efforts of large-scale data collection paint a bright future for the language.\\
\textbf{Kyrgyz and Turkmen Languages.} The current situation for Kyrgyz and Turkmen requires significant efforts for data collection and aggregation first. With the amount of data available for both languages, it seems barely useful not only to pre-train models like BERT, but also to research potential applications of statistical algorithms. In parallel with data collection, studies on transfer learning from Kazakh or Turkish might prove beneficial, given the grammatical similarity between the corresponding language pairs.\\
Additionally, given the incorporation of LLMs in the field, potential of using them for data augmentation or annotation can be another important direction of research in comparison with more expensive methods of human data collection and annotation.

\section{Conclusion}
Significant improvements in the processing of Turkic Central Asian languages have been achieved in the recent years. However, there still exists an imbalance in the amount of available data and technology, with Kazakh and Uzbek dominating the scene, and Kyrgyz and Turkmen requiring significant efforts in terms of data collection. Besides leveraging already existing datasets or web-sources for curating new data using data augmentation, other potentially successful options for technological advancements include transfer learning from Kazakh to Uzbek, Kyrgyz, and Turkmen. Already existing data in Kazakh might prove useful for the studies of other Central Asian languages, given their close linguistic relatedness.\\
While significant efforts are yet to emerge and be maintained for Kazakh, Uzbek, Kyrgyz, and Turkmen to reach the level of winners or underdogs \citep{joshi-etal-2020-state}, a substantial work has already been developed in the most crucial areas of natural language processing, including the automatic speech recognition systems and machine translation. With the potential of transferring this experience to other tasks, such as information retrieval, text generation, and question answering, there is a greater hope for progress towards the state-of-the-art technology for these Central Asian languages.

\section{Limitations}
We acknowledge several factors that limit the findings and scope of the presented survey. Firstly, we note that the rapid evolution of NLP technologies and data makes any work aimed at assessing current data sources and research developments outdated fairly quickly. Additionally, some previously published resources might become unavailable with time, which also contributes to the outdatedness. 
Secondly, we base the resource review process on the information provided by the authors of the relevant papers, which, however, might not reflect the real state of a dataset or a technology. There might exist some discrepancies between the data reported in the papers and those actually available due to access limitations, dataset updates, etc. Future work might address the above-mentioned limitations by establishing an open-source up-to-date list of resources and assessing the datasets quality empirically. 

\bibliography{coling_latex}

\clearpage

\appendix
\onecolumn

\section{Search Strategy}
\label{appendix:search}
\subsection{General Approach}
Firstly, we queried for papers using generic search terms and engines specified in \ref{appendix:search-engines}. Secondly, we focused on exploring popular CL and NLP venues, specific institutes' resources (e.g. ISSAI's website) as well as lists of references of works found in the first step of the process. Finally, we queried for pre-prints and unpublished works and datasets on arXiv, Github, Kaggle, and HuggingFace.

\subsection{Venues and Repositories}
The primary focus of our search was popular CL and NLP venues, including ACL, EACL, CoNLL, EMNLP, and WMT. We also explored non-ACL events and proceedings, including COLING and LREC.\\
For speech-related technology and datasets we researched works published at IEEE venues, including Interspeech, ICASSP, and SLT.\\
Additionally, we browsed pages of the universities and institutes that have been known for their contributions to the field, including ISSAI, Leipzig University, and Saarland University.\\
For pre-prints, works in progress, models, and datasets, we explored Github, Kaggle, and HuggingFace as well as the "Computation and Language" section on arXiv.

\subsection{Search Engines and Query Details}
\label{appendix:search-engines}
Search engines used include Google Scholar, Semantic Scholar, and ResearchGate. We used both broad and specific query terms to search for publications on both resources and technologies. In terms of broad queries, we used the format of "[language] [topic]", for example, "Kyrgyz NLP" or "Kazakh speech". For more technology-specific searches we adopted queries of the form of "[language] [technology name/data]", for example, "Kazakh NER" or "Uzbek speech corpus". \\
The search cutoff date is set to November 1, 2024.

\clearpage

\section{Datasets}
\label{appendix:datasets}
Below we provide statistics on the datasets surveyed in the main body of the paper. Datasets marked with an asterisk ("*") have not been released and/or are not currently available on an open-source basis. For the datasets marked with a dash ("-") in the last column, exact data quantities have not been reported. The amount of data provided is cited according to the datasets' authors report, which might differ from the actual data available.

\begin{longtable}[H]{ l p{5.5cm} p{4cm} p{4.5cm} }
\hline
\textbf{Language} & \textbf{Dataset} & \textbf{Type/Task} & \textbf{Data Available} \\
\hline
\endfirsthead

\hline
\textbf{Language} & \textbf{Dataset} & \textbf{Type/Task} & \textbf{Data Available} \\
\hline
\endhead

\hline
\endfoot

\textbf{Kazakh} & KazEmoTTS & sentiment analysis & 54.7K audio-text pairs \newline 74H \newline 8.7K unique sentences \\
\cline{2-4}
& KazSAnDRA & sentiment analysis & 180K \\
\cline{2-4}
& KazakhTTS & text-to-speech & 93H \\
\cline{2-4}
& KazakhTTS2 & text-to-speech & 271.7H \\
\cline{2-4}
& KSC2 & speech corpus & 1,200H \newline 600K utterances \\
\cline{2-4}
& Kazakh Speech Commands Dataset & speech commands & 119 speakers \newline >100K utterances \\
\cline{2-4}
& KOHTD & hand-written dataset & 3K exam papers \newline 140K segmented images \newline 922K symbols \\
\cline{2-4}
& KazNERD & NER & 25 entity classes \newline 112K sentences \newline 136K annotations \\
\cline{2-4}
& KazQAD & QA & 6K questions \newline 12K passage level judgments \\
\cline{2-4}
& Almaty Corpus (NCKL) & linguistic features & 40M word tokens \newline 650K words \\
\cline{2-4}
& Kazakh Language Corpus* & linguistic features & 135M words \newline 400K words \\
\cline{2-4}
& Kazakh KTB & universal dependencies & 300 sentences \\
\cline{2-4}
& kkwac* & corpora collection & 139M words \\
\cline{2-4}
& Leipzig Corpora & corpora collection & 51.4K news \newline 17M web \newline 773K Wikipedia sentences \\
\cline{2-4}
& Kazakh National Language Corpus & corpora collection & 22K docs\newline 23M words \\
\cline{2-4}
& Common Voice & speech corpus & 4H recorded \newline 3H validated \\
\hline
\textbf{Uzbek} & Restaurant Reviews \newline \cite{matlatipov2022uzbek} & sentiment analysis & 4.5K positive \newline 3.1K negative \\
\cline{2-4}
& Application Reviews \newline \cite{Kuriyozov2019BuildingAN} & sentiment analysis & 2.5K positive \newline 1.8K negative \\
\cline{2-4}
& Uzbek POS* \newline \cite{Sharipov2023UzbekTaggerTR} & POS & - 
\\
\cline{2-4}
& Multi-label Text Classification \newline
\cite{kuriyozov2023text} & classification & 512K articles \newline 120M words \newline 15 classes \\
\cline{2-4}
& Multi-label News Classification* \newline
\cite{Rabbimov2020MultiClassTC} & classification & 13K articles \\
\cline{2-4}
& Uzbek Speech Corpus & speech corpus & 105H \\
\cline{2-4}
& Common Voice & speech corpus & 265H recorded \newline 100H validated \\
\cline{2-4}
& UzWordNet & WordNet & 28K synsets \\
\cline{2-4}
& SimRelUz & linguistic features & 1.4K word pairs \\
\cline{2-4}
& Uzbek Electronic Corpus & corpora collection & - \\
\hline
& Leipzig Corpora & corpora collection & 86K news \newline 663K community \newline 280K newscrawl \newline 263K Wikipedia sentences \\
\hline
\textbf{Kyrgyz} & kloop & corpora collection & 16.8K articles \\
\cline{2-4}
 & kkwyc* & corpora collection & 19M words \\
\cline{2-4}
 & Manas-UdS & corpora collection & 1.2M words \\
\cline{2-4}
& Leipzig Corpora & corpora collection & 251K community \newline 123k newscrawl \newline 1.5K news \newline 3M web \newline 334K Wikipedia sentences\\
\cline{2-4}
 & Kyrgyz MNIST & hand-written symbols & 80K images \\
\cline{2-4}
 & UD-Kyrgyz-KTMU & universal dependencies & 781 sentences \newline 7.4K tokens \\
\cline{2-4}
 & Kyrgyz news dataset & classification & 23K articles \newline 20 classes \\
\cline{2-4}
 & Common Voice & speech corpus & 48H recorded \newline 39H validated \\
 \hline
\textbf{Turkmen} & Common Voice & speech corpus & 7H recorded \newline 3H validated \\
\cline{2-4}
& Leipzig Corpora & corpora collection & 276K web sentences \newline 62K Wikipedia sentences \\
\hline
\end{longtable}

\captionof{table}{Overview of monolingual datasets and their availability per language.}
\label{table:mono-datasets}

\clearpage

\begin{longtable}[t]{ p{4cm} c p{4.5cm} p{4.5cm} }
\hline
\textbf{Dataset} & \textbf{Languages} & \textbf{Type/Task} & \textbf{Data Available} \\
\hline
\endfirsthead

\hline
\textbf{Dataset} & \textbf{Languages} & \textbf{Type/Task} & \textbf{Data Available} \\
\hline
\endhead

\hline
\endfoot

KazParC & KK, RU, EN, TR & parallel corpora & 3K exam papers \newline 140K segmented images \newline 922K symbols \\
\hline
Russian-Kazakh \newline Handwritten Database & RU, KZ & handwritten symbols & 63K sentences \newline 95\% RU, 5\% KZ\\
\hline
KRSL* & KK, RU & sign language & 890H of videos \newline 325 annotated videos \newline 39K gloss annotations \\
\hline
Uzbek-Kazakh Corpora & UZ, KK & parallel corpora & 124K sentences \\
\hline
Large Turkic Language Corpora* \newline \cite{Baisa2012LargeCF} & KK, KY, UZ, TK & web corpora collection \newline morphological segmentation & 1.4M KZ \newline 590K KY \newline 320K UZ \newline 200K TR words\\
\hline
Belebele & KK, UZ, KY & reading comprehension & 900 questions \newline 488 passages \\
\hline
xSID & KK & syntactic data & - \\
\hline
CC100 & KK, KY, UZ & web corpora collection & - \\
\hline
WikiAnn & KK, KY, UZ, TK & NER & - \\
\hline
OSCAR & KK, KY, UZ, TK & web corpora collection & 677K KK \newline 144K KY \newline 15K UZ \newline 4.5K TR docs \\
\hline
AM2iCO & KK & lexical alignment & - \\
\hline
ST-kk-ru & KK, RU & speech translation & 317H \\
\hline
M2ASR* & KK, KY & speech recognition & - \\
\hline
MuMiN & KK & multimodal fact checking & - \\
\hline
\end{longtable}

\captionof{table}{Overview of multilingual and/or parallel datasets and their availability per language.}
\label{table:multi-datasets}

\end{document}